\def\DatasetName{Koala-36M~}

\documentclass[10pt,twocolumn,letterpaper]{article}

\usepackage{cvpr}              % To produce the CAMERA-READY version
\usepackage{booktabs}
\usepackage{makecell}
\definecolor{cvprblue}{rgb}{0.21,0.49,0.74}
\usepackage[pagebackref,breaklinks,colorlinks,allcolors=cvprblue]{hyperref}

\def\paperID{2517} % *** Enter the Paper ID here
\def\confName{CVPR}
\def\confYear{2025}

\title{\DatasetName: A Large-scale Video Dataset \\Improving Consistency between Fine-grained Conditions and Video Content}

\author{
Qiuheng Wang$^{1,3}$ \thanks{Equal contribution. Work done as an intern at Kuaishou Technology.~\qquad$^{\dag}$ Corresponding author.} ~\qquad Yukai Shi$^{2,3}$\footnote[1]~\qquad
Jiarong Ou$^{3}$~\qquad
Rui Chen$^{3}$~\qquad
Ke Lin$^{3}$~\qquad
\\
Jiahao Wang$^{3}$~\qquad
Boyuan Jiang$^{3}$~\qquad
Haotian Yang$^{3}$~\qquad
Mingwu Zheng$^{3}$~\qquad
\\
Xin Tao$^{3}$~\qquad
Fei Yang$^{3}$$^{\dag}$~\qquad
Pengfei Wan$^{3}$~\qquad
Di Zhang$^{3}$\\
$^1$~Shenzhen University~\qquad
$^2$~Tsinghua University~\qquad
$^3$~Kuaishou Technology
}
\vspace{-0.5em}
\begin{document}

\twocolumn[{%
    \renewcommand\twocolumn[1][]{#1}%
    \maketitle
    \vspace{-3em}
    \begin{center}
    \includegraphics[width=\textwidth]{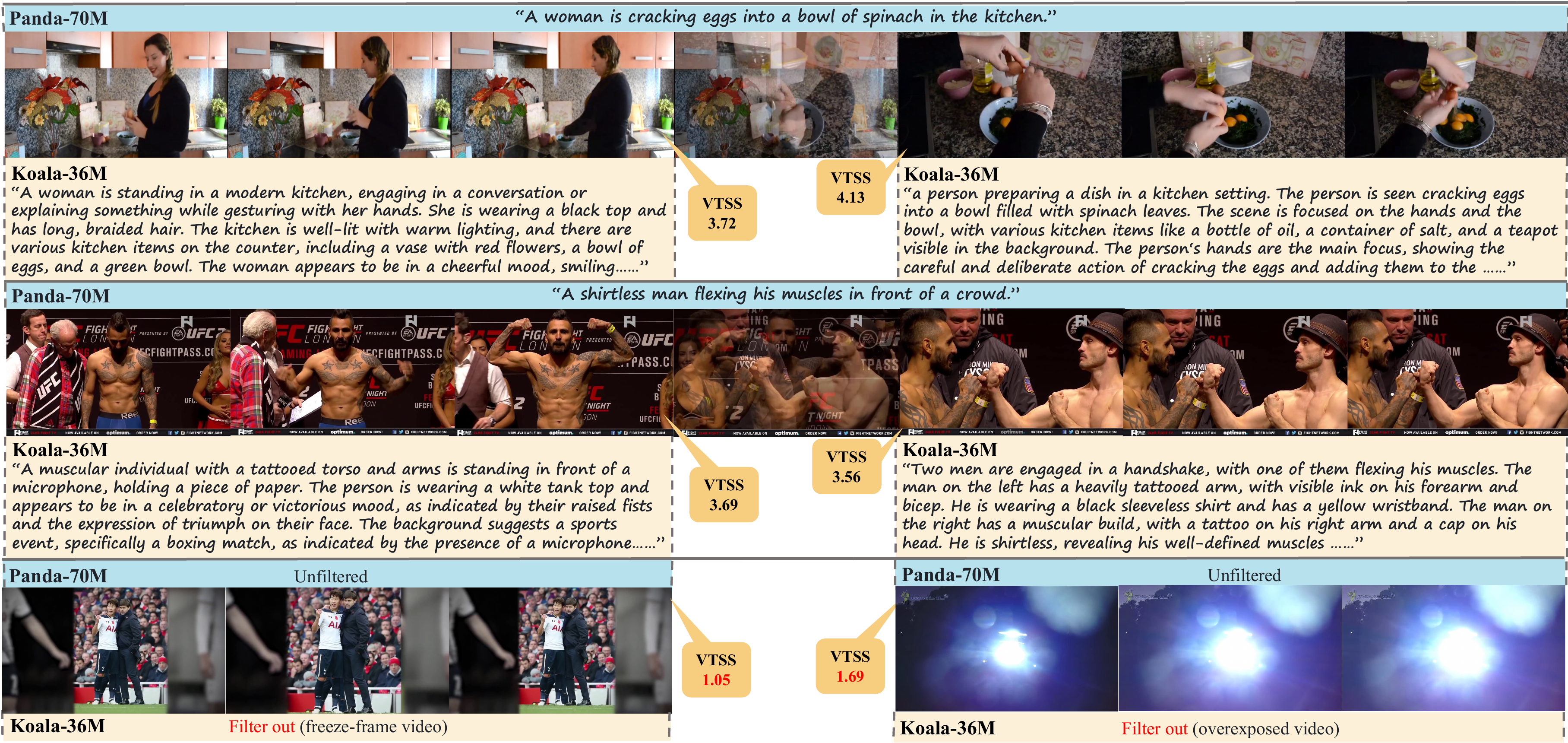}
    \end{center}
    \vspace{-1em}
    \captionof{figure}{\textbf{Comparison between \DatasetName and Panda-70M.} We propose a large-scale, high-quality dataset that significantly enhances the consistency between multiple conditions and video content. \DatasetName features more accurate temporal splitting, more detailed captions, and improved video filtering based on the proposed Video Training Suitability Score (VTSS).}
    \label{fig:showcase}
    \vspace{1em}
}]

\renewcommand{\thefootnote}{\fnsymbol{footnote}}
\footnotetext[1]{Equal contribution. $^{\dag}$Corresponding author.}

\begin{abstract}
\vspace{-2.1em}

With the continuous progress of visual generation technologies, the scale of video datasets has grown exponentially. The quality of these datasets plays a pivotal role in the performance of video generation models. We assert that temporal splitting, detailed captions, and video quality filtering are  three crucial determinants of dataset quality. However, existing datasets exhibit various limitations in these areas. To address these challenges, we introduce \textbf{\DatasetName}, a large-scale, high-quality video dataset featuring accurate temporal splitting, detailed captions, and superior video quality. The essence  of our approach lies in improving the consistency between fine-grained conditions and video content. Specifically, we employ a linear classifier on probability distributions to enhance the accuracy of transition detection, ensuring better temporal consistency. We then provide structured captions for the splitted videos, with an average length of 200 words, to improve text-video alignment. Additionally, we develop a \textit{Video Training Suitability Score (VTSS)} that integrates multiple sub-metrics, allowing us to filter high-quality videos from the original corpus. Finally, we incorporate several metrics into the training process of the generation model, further refining the fine-grained conditions. Our experiments demonstrate the effectiveness of our data processing pipeline and the quality of the proposed \DatasetName dataset. Our dataset and code have been released at {\url{https://koala36m.github.io/}}.
\end{abstract}

\vspace{-2.2em}
\section{Introduction}
\vspace{-1em}

Generative AI, especially in video generation tasks, has recently captivated the attention of researchers. These tasks entail generating high - quality videos from textual descriptions or images. The quality of the datasets used for training is a decisive factor in the success of these models. Multiple open - source datasets, such as Panda - 70M\citep{chen2024panda}, MiraData~\citep{ju2024miradata}, OpenVid~\citep{nan2024openvid}, and VidGen~\citep{tan2024vidgen}, have been introduced. Each of these datasets carefully selects data sources and applies diverse evaluation metrics for video filtering. Additionally, innovative methods, like the multi - modal caption model~\citep{chen2024panda} or structured captions~\citep{ju2024miradata}, have been utilized in the video captioning process.

Despite the success of the data processing pipelines introduced by previous datasets, we argue that the core challenge lies in establishing accurate and fine - grained conditioning for video data, which is crucial for both reducing the complexity of the training process and improving the quality of the generated outputs. To achieve this, we believe there are three key issues that need to be addressed:

First, the alignment between text and video semantics is essential. Unlike video question - answering tasks, where captions are primarily driven by specific question - based details, video generation requires captions that are directly tied to the visual content itself. Given the infinite granularity of visual signals, this demands captions that are rich and detailed. Moreover, raw video data often contains complex transitions, posing additional challenges in ensuring caption accuracy.
Second, the effective evaluation and filtering of low - quality data remain underexplored. Low - quality video data, such as those with poor visual quality or excessive artificial effects, can impede the training process. However, accurately assessing and filtering such data presents an ongoing challenge. Existing methods typically rely on manually selected quality metrics and heuristic threshold - based filtering, which are often designed for other tasks and may not meet the specific requirements of video generation. Consequently, these approaches may not effectively guarantee the desired data quality for training.

Third, even with data filtering processes in place, the videos within the dataset still vary in quality, with each video potentially having different strengths and weaknesses (e.g., one video may have lower clarity but better aesthetic appeal). Training with such heterogeneous data in the same way may introduce ambiguity for the model, hampering its ability to learn effectively.

To address these issues, we present \textbf{\DatasetName}, a large-scale high-quality video dataset with more accurate video splitting, detailed captions, better data filtering methods and metric conditions. 
As video content reaches considerable quality, the consistency between fine-grained conditions and video content determines the performance of generation models. Based on this crucial understanding, we propose a more sophisticated data processing pipeline.
Since accurate video splitting leads to better temporal consistency, we first employ a linear classifier on probability distributions to enhance the accuracy of transition detection.
Then we generate structured captions for the segmented  video clips, with an average length of 200 words, to improve text-video alignment.
Moreover, to prevent the accidental elimination of high-quality data during the filtering procedure, we train a network on human - curated datasets to predict the \textit{Video Training Suitability Score (VTSS)}. This network models the joint distribution of sub-metrics by taking videos and sub - metrics as inputs and outputs a single \textit{Video Training Suitability Score}, which serves as the sole metric for data filtering.
Additionally, we introduce data metrics as supplementary  conditions (\textit{Metric Conditions}) into the generation model during training. This enables the model to differentiate data of varying quality levels and further enhances the consistency between fine - grained conditions and video content, which results in better performance and controllability of the generation model.

To further validate the efficacy of \DatasetName and our data processing pipeline, we conduct training of video generation models on different datasets. Both the dataset benchmark and the performance of the video generation model demonstrate the advantage of the \DatasetName dataset. We perform more ablation studies to demonstrate effectiveness of our data processing pipeline.

% \vspace{-0.2em}
Our contributions can be summarized as follows:
% \vspace{-0.7em}
\begin{itemize}
    \item We present a large-scale high-quality dataset called \DatasetName, with accurate video splitting, detailed captions and higher-quality video content.
    % % \vspace{-0.3em}
    \item We propose a refined data processing pipeline to further improve the consistency between fine-grained conditions and video content, including transition detection methods, structured caption system, Video Training Suitability Score and metric conditions.
    % \vspace{-0.5em}
    \item Comprehensive experiments demonstrate the advantages of \DatasetName dataset and the effectiveness of our data processing pipeline.
\end{itemize}

\vspace{-0.5em}

\section{Related Work}
\vspace{-0.5em}
Recent progress in diffusion models has spurred the evolution of image generation models into video generation models. In the realm of text-to-video (T2V) generation, considerable efforts have been exerted to develop large-scale T2V models. These models are trained on extensive datasets, making use of traditional U-Net-based diffusion architectures~\citep{zeng2024make, clark2024text, ge2023preserve, yu2023video, khachatryan2023text2video} and Transformer - based (DiT) architectures~\citep{ma2024latte, chen2023gentron, lu2023vdt, chen2024gentron, xing2024simda}. The success of these video generation models is critically reliant on the quality of the video-text datasets.

\vspace{-0.5em}
\subsection{Video datasets}
\vspace{-0.5em}
While several video datasets~\citep{caba2015activitynet,anne2017localizing,rohrbach2015dataset,zhou2018towards,xu2016msr,internvid,how2,videofactory, chen2023videocrafter1} have been applied to tasks such as action recognition, video understanding, visual question answering (VQA), and video retrieval, there remains an urgent need for a high-quality, open-source dataset specifically tailored for training video generation models, providing rich video-text pairs. Datasets such as YouCook2~\citep{zhou2018towards}, VATEX~\citep{wang2019vatex}, and ActivityNet~\citep{caba2015activitynet} offer high-quality human caption annotations. Another set of datasets, including Miradata~\citep{ju2024miradata}, VidGen-1M~\citep{tan2024vidgen}, and OpenVid-1M~\citep{nan2024openvid}, automatically generate high-quality captions and filter data using manually selected thresholds on multiple dataset metrics. 

However, these datasets are insufficient in size to support the training of large models. Datasets, including YT-Temporal-180M~\citep{zellers2021merlot}, HD-VILA-100M~\citep{xue2022advancing}, ACAV~\citep{lee2021acav100m}, etc., contain hundreds of millions of video-text pairs, but their captions are automatically generated via speech recognition, leading to subpar quality. Panda70M~\citep{chen2024panda}, the largest publicly available video - text dataset, has gained popularity for video generation tasks owing to its scale and relatively good quality. Nevertheless, its quality still demands further enhancement. In particular, the captions in Panda - 70M frequently offer simplistic and incomplete descriptions of video content. Moreover, the frequent transitions in the training videos can cause semantic incoherence, potentially giving rise to undesirable or unpredictable transitions in the generated videos.

\vspace{-0.6em}

% table
\begin{table}[htbp]
    \centering
    \footnotesize
    \caption{\textbf{Comparison of \DatasetName and pervious text-video datasets.} \DatasetName is a video dataset that simultaneously possesses a large number of videos (over 10M) and high-quality fine-grained captions (over 200 words). We propose \textit{structured captions} and \textit{an expert model} (Video Training Suitability Score) for accurate data filtering. "TVL" and "ATL" are abbreviations for "Total Video Length" and "Average Text Length".}
    \vspace{-0.5em}
    \label{tab:related_work_dataset}
    \setlength{\tabcolsep}{0mm}
    \resizebox{\linewidth}{!}{
    \begin{tabular}{lccccccccc}
    \toprule
    Dataset & \#Videos  & ATL(words)  & TVL(hours) & Text & Filtering & Resolution \\
    \midrule
    LSMDC~\citep{rohrbach2015dataset}   & 118K  & 7.0    & 158       & Manual    & Sub-metrics     & 1080p \\
    DiDeMo~\citep{anne2017localizing}  & 27K  & 8.0     & 87         & Manual    & Sub-metrics    & -     \\
    YouCook2~\citep{zhou2018towards} & 14K & 8.8   & 176    & Manual    & Sub-metrics       & -     \\
    ActivityNet~\citep{caba2015activitynet} & 100K & 13.5   & 849  & Manual    & Sub-metrics    & -     \\
    MSR-VTT~\citep{xu2016msr} &  10K  & 9.3    & 40       & Manual    & Sub-metrics       & 240p  \\
    VATEX~\citep{wang2019vatex} & 41K  & 15.2   & $\sim$115      & Manual    & Sub-metrics    & - \\
    WebVid-10M~\citep{bain2021frozen}  &  10M & 12.0   & 52K    & Alt-Text    & Sub-metrics  & 360p     \\
    HowTo100M~\citep{howto100m} & 136M & 4.0     & 135K     & ASR     & Sub-metrics  & 240p  \\
    HD-VILA-100M~\citep{xue2022advancing} & 103M & 17.6    & 760.3K  & ASR   & Sub-metrics   & 720p  \\

    VidGen~\citep{tan2024vidgen} & 1M & 89.3    & -   & Generated & Sub-metrics     & 720p \\
    MiraData~\citep{ju2024miradata}  &  330K & 318.0    &  16K  &  Generated \& Struct & Sub-metrics    & 720p \\
    Panda-70M~\citep{chen2024panda} & 70M & 13.2    & 167K   & Generated & Sub-metrics     & 720p \\
    \midrule
    \textbf{\DatasetName(Ours)}  & 36M & 202.1    &  172K  &  Generated \& Struct & Expert Model    & 720p \\
    \bottomrule
    \end{tabular}
    }
\end{table}
\vspace{-0.8em}

\vspace{-0.5em}
\subsection{Video data curation}
\vspace{-0.5em}
As models continuously expand in scale, effective data curation is of paramount importance~\citep{zhou2023dataset}, particularly in the formulation of a well-suited training dataset. This is crucial for enhancing model performance and improving training efficiency during both the pretraining and supervised fine-tuning phases.
In the realm of large language models (LLMs), various data curation approaches have been proposed~\citep{xie2023data, maharana2023d2, tirumala2023d4}, including optimizations for data quantity, data quality, and domain composition. However, research on exploring data curation strategies within the video domain remains scarce. Stable Video Diffusion~\citep{blattmann2023stable} offers a comprehensive review of the curation of large - scale video datasets, encompassing techniques such as video clipping, captioning, and filtering.  Regrettably, the dataset is not open-source. In this study, we propose a novel data processing pipeline for video data and introduce a new video filtering metric. Distinct from traditional video quality assessment models~\citep{wu2023exploring, zhao2023zoom, wu2022fast, sun2024enhancing}, which mainly concentrate on the aesthetic and technical aspects of a video, our approach places emphasis on the suitability of videos for training purposes.

\vspace{-0.5em}
\section{\DatasetName Dataset}

\vspace{-0.5em}

\begin{figure}[h]
\begin{center}
\includegraphics[scale=0.4]{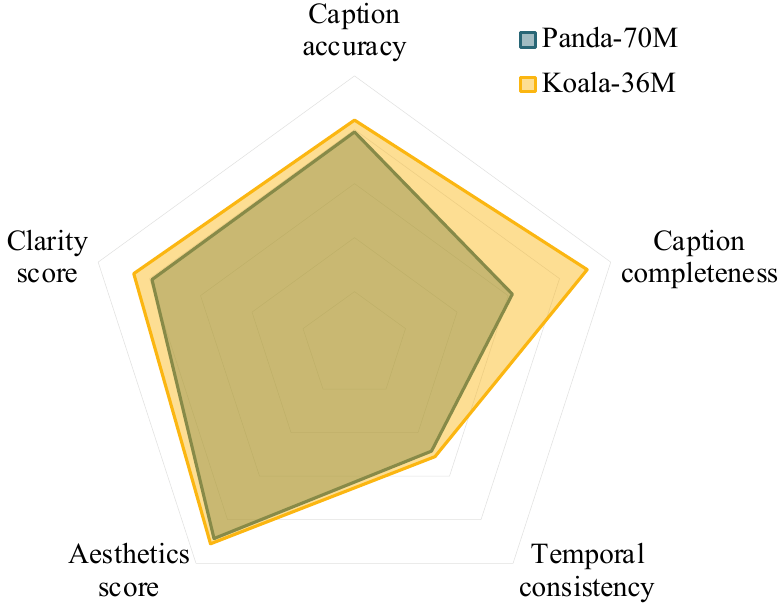}
\end{center}
\vspace{-1em}
\caption{\textbf{Quantitative comparison with Panda-70M.} \DatasetName has a significant improvement in the consistency between fine-grained conditions and video content.}
\label{fig:overview-lidar}
\end{figure}

\DatasetName is a large-scale high-quality video dataset with accurate video splitting, detailed captions and higher-quality video content. In summary, \DatasetName contains 36 million video clips with an average duration of 13.75 seconds and a resolution of 720p, each captioned by a text description averaging 202 words in length. We compare \DatasetName dataset with previous video datasets in Tab.~\ref{tab:related_work_dataset}. \DatasetName dataset simultaneously provides a large number of videos (over 10M) and high-quality fine-grained text captions (longer than 200 words), significantly improving the quality of large scale video datasets. Additionally, as shown in Fig.~\ref{fig:overview-lidar}, we further compare \DatasetName with Panda-70M on a series of dataset metrics, such as aesthetic scores and clarity scores, demonstrating a significant improvement in consistency between fine-grained conditions and video content. Since these two datasets come from the same raw datasets, the superiority of Koala-36M dataset also prove the effectiveness of our data processing pipeline.

\vspace{-0.3em}

\vspace{-1.5em}
\vspace{-0.5em}
\section{Method}
\begin{figure*}[htbp]
\begin{center}
\includegraphics[scale=0.3]{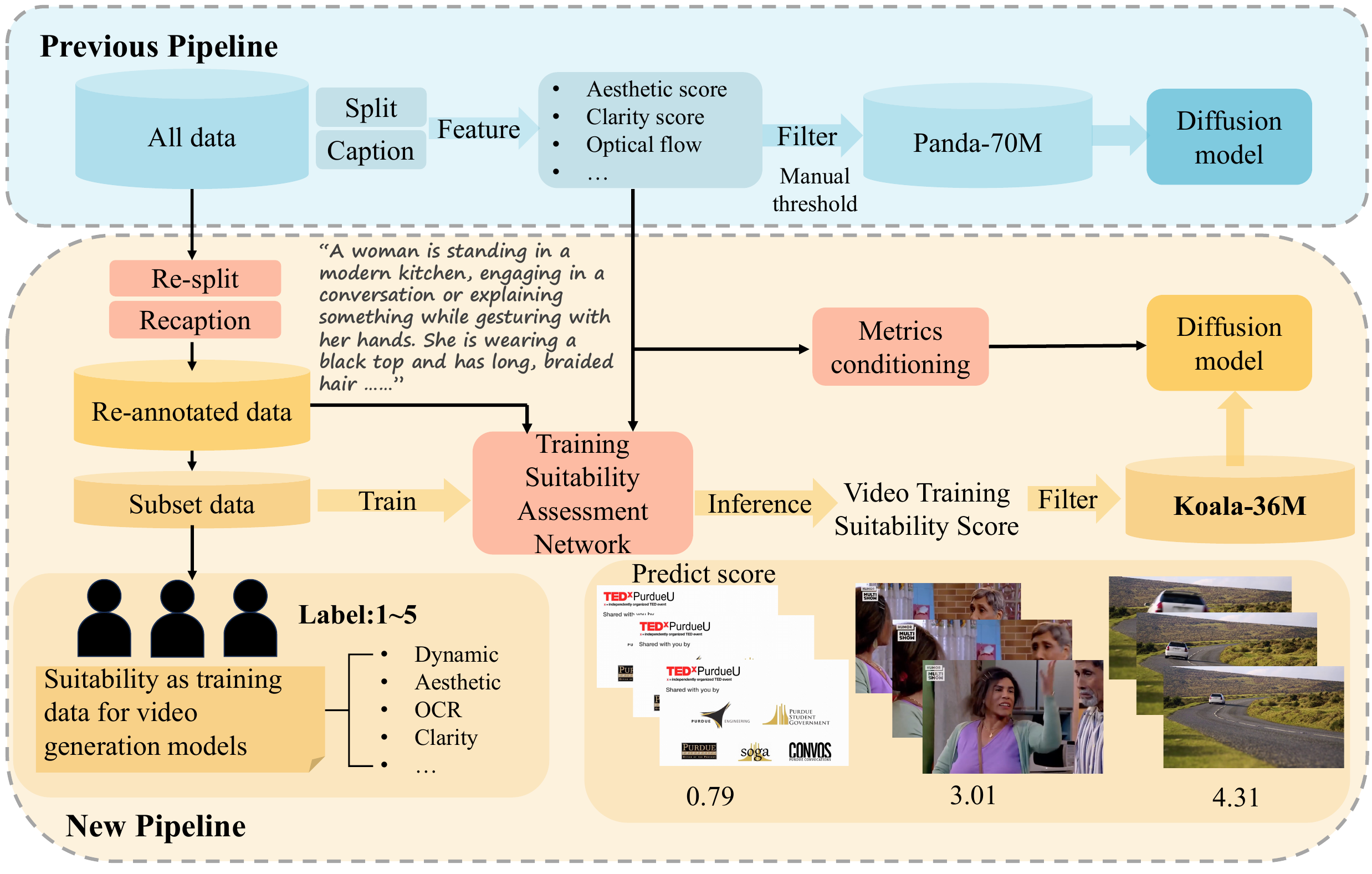}
\end{center}
\vspace{-1.5em}
\caption{\textbf{The proposed data processing pipeline.} Compared with previous pipeline, we propose better splitting methods, structured caption system, training suitability assessment network and metrics conditioning in red box, improving the consistency between conditions and video content.}
\vspace{-1.5em}
\label{fig:pipeline}
\end{figure*}

\vspace{-0.5em}
As shown in Fig.~\ref{fig:pipeline}, we propose a refined data processing pipeline for \DatasetName dataset. Our pipeline aims to further improve the consistency between fine-grained conditions and video content. Our main contributions are highlighted  in the red box of Fig.~\ref{fig:pipeline}. Specifically, we start from the same raw data with Panda-70M~\citep{chen2024panda} dataset. Firstly, in section~\ref{section: video splitting}, we propose a more precise and efficient transition detection approach for video segmentation. Subsequently, in section~\ref{section: video caption}, we generate captions for the segmented videos, with an average length of 200 words, using our structured caption system. Next, in section~\ref{section: data filtering}, we train a Video Training Suitability Score (VTSS) for data filtering, aiming to prevent the inadvertent deletion of high - quality data. Finally, in section~\ref{section: comprehensive perception conditions}, we incorporate multiple data sub - metrics as \textit{Metric Conditions} into the generation model to enrich the fine - grained conditions.

\vspace{-0.5em}
\subsection{Video splitting}
\label{section: video splitting}

\vspace{-0.5em}
Splitting videos into temporal segments is crucial for creating video generation datasets. Transition-free video data enable more accurate alignment between text and video, while reducing the difficulty of model training and improving the temporal consistency of generated results. Existing  video splitting techniques  Pyscenedetect~\citep{pyscene} typically detect transitions based on changes in image features between consecutive frames, relying on manually adjusted thresholds as criteria, but often overlook temporal information. As a result, these methods struggle to distinguish between gradual transitions and fast-motion scenes, leading to missed detections in the former and incorrect detections in the latter.

To tackle the aforementioned problems, we initially introduce a Color-Struct SVM (CSS) module. This module employs a learning-based strategy, enabling more precise detection of frame-to-frame changes compared to threshold-based methods. Subsequently, we utilize temporal smoothing and statistical features to distinguish between gradual transitions and fast-motion scenes.

We assume that transitions occur with a low probability at any given moment in the video. We treat image pairs from the same video source as negative examples and pairs from different video sources as positive examples. We  select BGR histogram correlation to measure color distance and Canny Luminance SSIM to measure structural distance, which together measure inter-frame changes. For images $ I_{i} $ and $ I_{j} $ , the color distance $d_{color}$ and structural distance $d_{struct}$ are defined as follows:

\begin{equation}
   \begin{aligned}
        H_i &= \text{Histogram}(bgr(I_{i}))
    \label{eq:d_all}
   \end{aligned}  
\end{equation}

\begin{equation}
   \begin{aligned}
   d_{color}(H_{i}, H_{j}) &= \frac{\sum_{p} (H_{i}(p) - \bar{H_{i}}) (H_{j}(p) - \bar{H_{j}})}{\sqrt{ \sum_{p} (H_{i}(p) - \bar{H_{i}})^2 (H_{j}(p) - \bar{H_{j}})^2 }}
   \label{eq:d_color}
   \end{aligned}
\end{equation}

\begin{equation}
    \begin{aligned}
        E_{i} &= \max(\text{Gray}(I_{i}), \text{Canny}(\text{Gray}(I_{i})))
        \label{eq:E_i}
    \end{aligned}
\end{equation}

\begin{equation}
    \begin{aligned}
        d_{struct}(E_{i}, E_{j}) &= \text{SSIM}(E_{i}, E_{j})
    \label{eq:d_struct}
    \end{aligned}
\end{equation}

Then an SVM classifier is employed, using color distance $d_{color}$ and structural distance $d_{struct}$ as the relevant input features; see Eq.~\ref{eq:d_all}, Eq.~\ref{eq:d_color}, Eq.~\ref{eq:E_i}, Eq.~\ref{eq:d_struct} . Regarding temporal information, we hypothesize that video changes are relatively stable over time. By estimating a Gaussian distribution of changes from past frames, if the current frame's change exceeds the $3\sigma$ confidence interval, we consider it a significant transition. This method enhances the differentiation between gradual transitions and fast-motion scenes without increasing computational load. Extensive experiments demonstrate the effectiveness of the transition detection method in ~\ref{app:effectiveness of video splitting methods}.

\vspace{-0.5em}
\subsection{Video captioning}
\label{section: video caption}
\vspace{-0.5em}
Detailed captions typically result in enhanced text - video consistency, which significantly influences the granularity of semantic responses. To obtain more detailed captions, we propose a structured caption system, which consists of: (1) the subject, (2) actions of the subject, (3) the environment in which the subject is located, (4) the visual language  including style, composition, lighting, etc. (5) the camera language including camera movement, angles, focal length, shot sizes, etc. (6) world knowledge. We generate these aspects separately, and merge them as the final caption.

\begin{figure}[h]
\begin{center}
\includegraphics[scale=0.4]{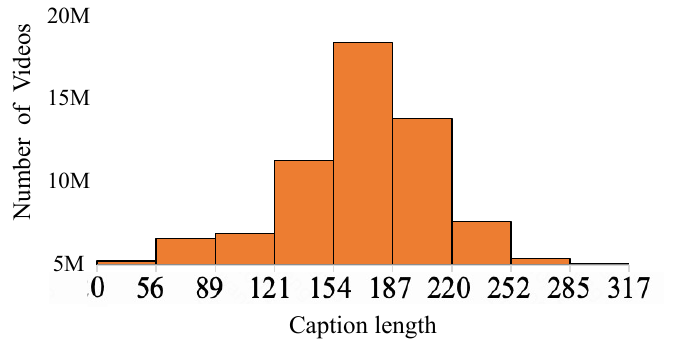}
\end{center}
\vspace{-1.5em}
\caption{\textbf{Distribution of the caption length} (in words) in \DatasetName dataset.}
\vspace{-1em}
\label{fig:caption distribution}
\end{figure}

Similar to previous works~\citep{chen2024panda,tan2024vidgen,ju2024miradata}, we first collect a caption dataset by using GPT-4V~\citep{gpt4v} to generate video captions based on our structured system. We then fine-tune a caption model based on LLaVA~\citep{liu2023llava} for the entire dataset. Our experiments during fine-tuning show that training the vision encoder improves the accuracy of the caption. And a high-resolution vision encoder helps the caption model capture video details better. To alleviate the computational burden caused by high-resolution inputs, we perform average pooling with a 2x2 kernel on the spatial dimensions of the tokens, ensuring minimal information loss. Notably, we implement  a mixed training strategy involving both static images and dynamic videos, enabling the model to concurrently learn visual understanding in both static and dynamic scenarios. This also enhances data diversity, alleviating the issue of insufficient training samples when solely relying on video data.

Finally, we run our captioner on the whole dataset, and the distribution of caption lengths is shown in Fig.~\ref{fig:caption distribution}. Furthermore, we evaluate the quality of captions with caption accuracy and completeness. As shown in Fig.~\ref{fig:overview-lidar} and Tab.~\ref{tab:related_work_dataset}, our structured caption system significantly improve the quality of captions with better text-video consistency.

% data filtering
\vspace{-0.3em}
\subsection{Data filtering}
\begin{figure}[h]
\begin{center}
\vspace{-0.5em}
\includegraphics[width=\linewidth]{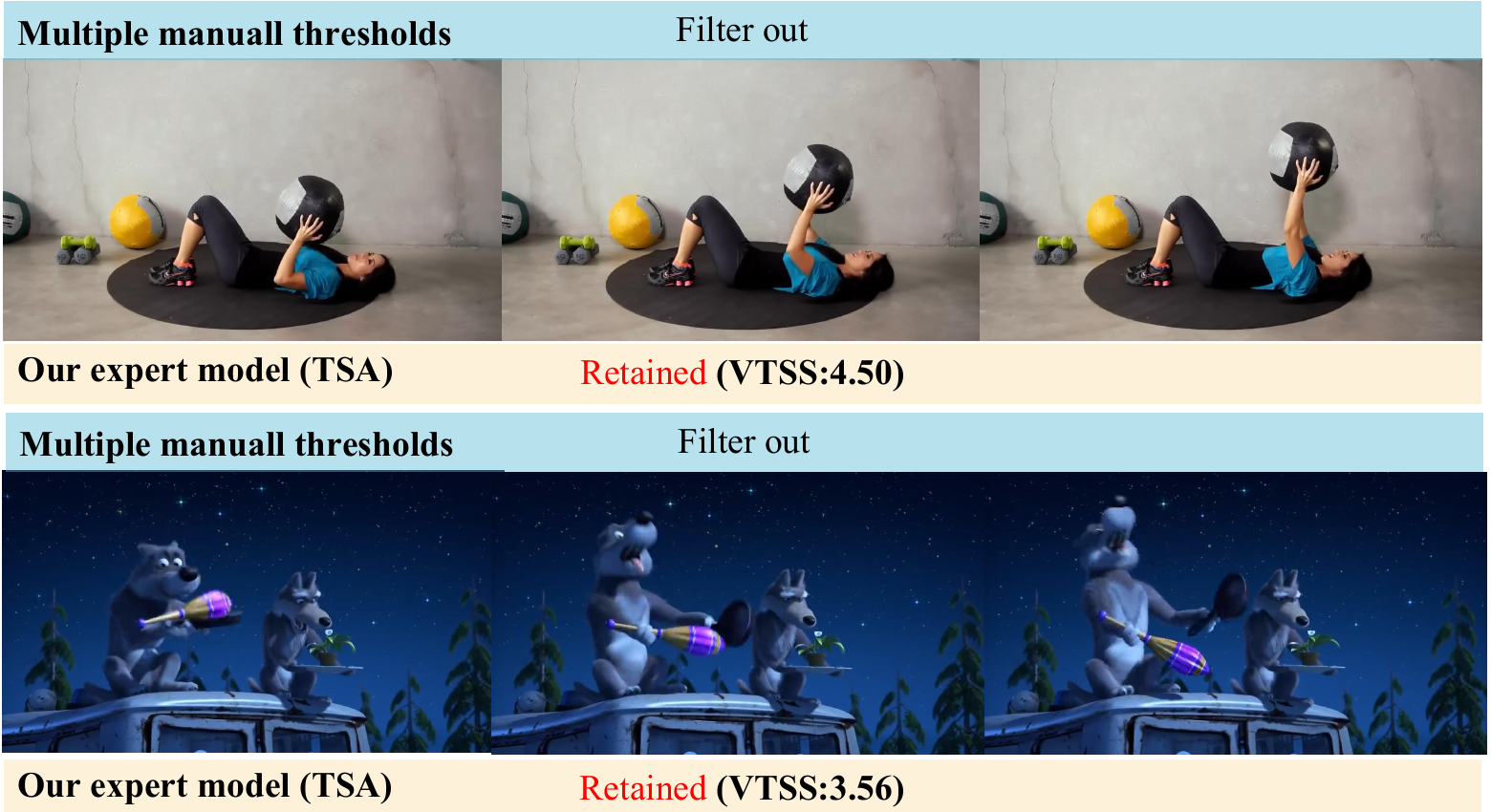}
\end{center}
\vspace{-1.5em}
\caption{\textbf{The deleted high-quality data by inaccurate multiple manual thresholds.}}
\vspace{-1em}
\label{fig:mistaken_deletion}
\end{figure}

\vspace{-0.3em}
\label{section: data filtering}
In the large-scale raw dataset, the quality of video content varies significantly. When the performance of the generation model hinges on videos with considerable content quality, it is both necessary and crucial to filter out low-quality data and retain high-quality data accurately. Traditional approaches typically utilize  various sub-metrics to evaluate video quality and then manually set thresholds to filter the desired data. Since these sub-metrics are not completely orthogonal with each other, the video quality is actually a joint distribution of all sub-metrics, which means these thresholds should have implicit constraints with each other. However, existing methods neglect the joint distribution of sub-metrics, resulting in inaccurate thresholds. Meanwhile, since multiple thresholds need to be set, the cumulative effect of inaccurate threshold leads to larger deviations during filtering. Therefore, not only low-quality videos are not correctly filtered out in Fig.~\ref{fig:showcase}, but also high-quality videos are mistakenly deleted in Fig.~\ref{fig:mistaken_deletion}. More analysis and experiments are shown in App.~\ref{app:data filtering analysis}.

To address this issue, we propose a \textit{Training Suitability Assessment Network (TSA)} to model the joint distribution of sub-metrics. This network takes videos and sub-metrics as input, and outputs a single value called \textit{Video Training Suitability Score (VTSS)} as the only metric to filter data. This score reflects whether a video is suitable for training purposes. Specifically, we first collect the training set from human evaluation based on a new criteria. Then we train the \textit{Training Suitability Assessment Network (TSA)} and employ it to calculate VTSS for all videos. Finally, we set a single threshold for VTSS based on its distribution for filtering.

\vspace{-0.3em}
\subsubsection{New criteria and human evaluation}
\vspace{-0.3em}
We have defined a new annotation criterion that assigns a score reflecting whether a video is suitable as training data for video generation models. This criterion chiefly takes into account the following dimensions of video quality:
\textbf{Dynamic Quality}: A high-quality video should exhibit good dynamics, which are evaluated based on two factors: the extent of subject movement and the temporal stability of the motion. The motion area in the video should cover more than 30\% of the frame; otherwise, the score of the video will be decreased for insufficient dynamics. Temporal stability pertains  the camera movement; non-professional videographers often produce videos with irregular and significant shaking. We decrease the scores of such videos to distinguish them from professional works.
\textbf{Static Quality}: Each frame of a high-quality video should have rich subject details, reasonable composition, aesthetic appeal, clear and distinct subjects, and vividly saturated colors. Although this metric may involve some subjectivity, it is crucial for assessing the overall visual quality.
\textbf{Video Naturalness}: We prefer videos that are natural and unprocessed. Special effects, transitions, subtitles, and logos can introduce biases in the original video distribution , making it harder for generation models to learn.  Additionally, we consider the safety of the video content, rejecting videos with political, terrorist, violent, gory, or otherwise disturbing content. In order to reduce the bias between the labeled scores and the true scores, each video is labeled by 8 experts and subjected to a bias elimination process in the App~\ref{app:label}.

\vspace{-0.3em}
\subsubsection{Training Suitability Assessment Network}
\vspace{-0.3em}

\begin{figure}[h]
\begin{center}
\includegraphics[scale=0.25]{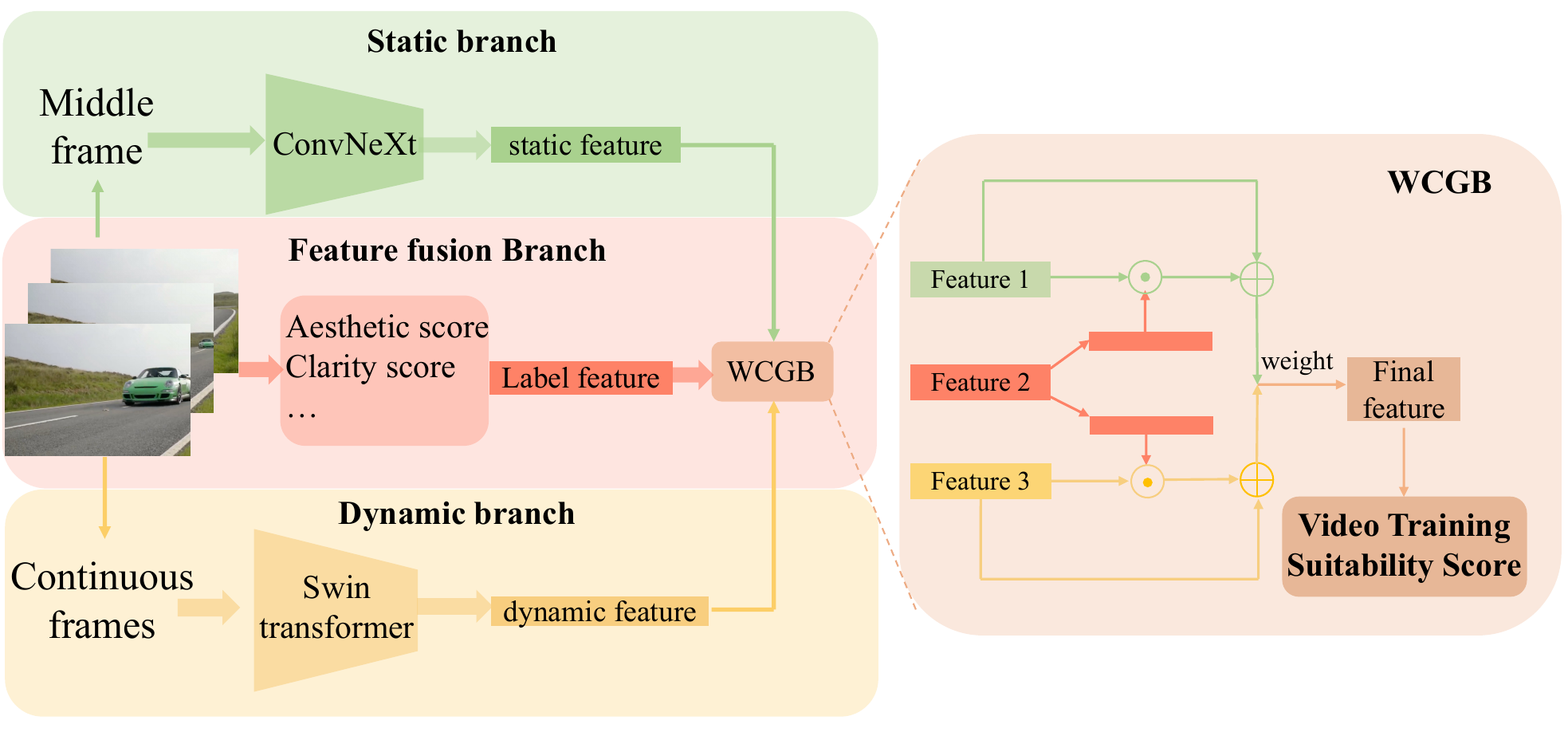}
\end{center}
\vspace{-1.5em}
\caption{\textbf{Training Suitability Assessment Network.}}
\vspace{-1.0em}
\label{fig:TSA pipeline}
\end{figure}

As shown in Fig.~\ref{fig:TSA pipeline}, we propose a \textit{Training Suitability Assessment Network}, which takes videos and sub-metrics as input, and outputs a single value called \textit{Video Training Suitability Score (VTSS)}.  Aligned with the previously mentioned annotation criteria, our network is structured into dynamic and static branches. Furthermore, we maintain diverse data labels from conventional data filtering approaches and introduce this supplementary information to the network model as an additional branch.
For the features of different branches, 3D Swin Transformer is employed  as the backbone for the dynamic branch, while the ConvNext network for the static branch. To integrate the features from different branches, we propose a Weight Cross-Gating Block (WCGB) to incorporate the information from the label branch into the other two branches.
Since the label branch inherently reflects various characteristics of the video, which are related to both dynamic and static features, we use label features to enhance the dynamic and static features. Given that different video labels focus on dynamic and static aspects to varying degrees, we learn a fusion weight to adjust the proportion of label features integrated with the two types of video features.

\begin{figure}[h]
\begin{center}
\vspace{-0.8em}
\includegraphics[scale=0.15]{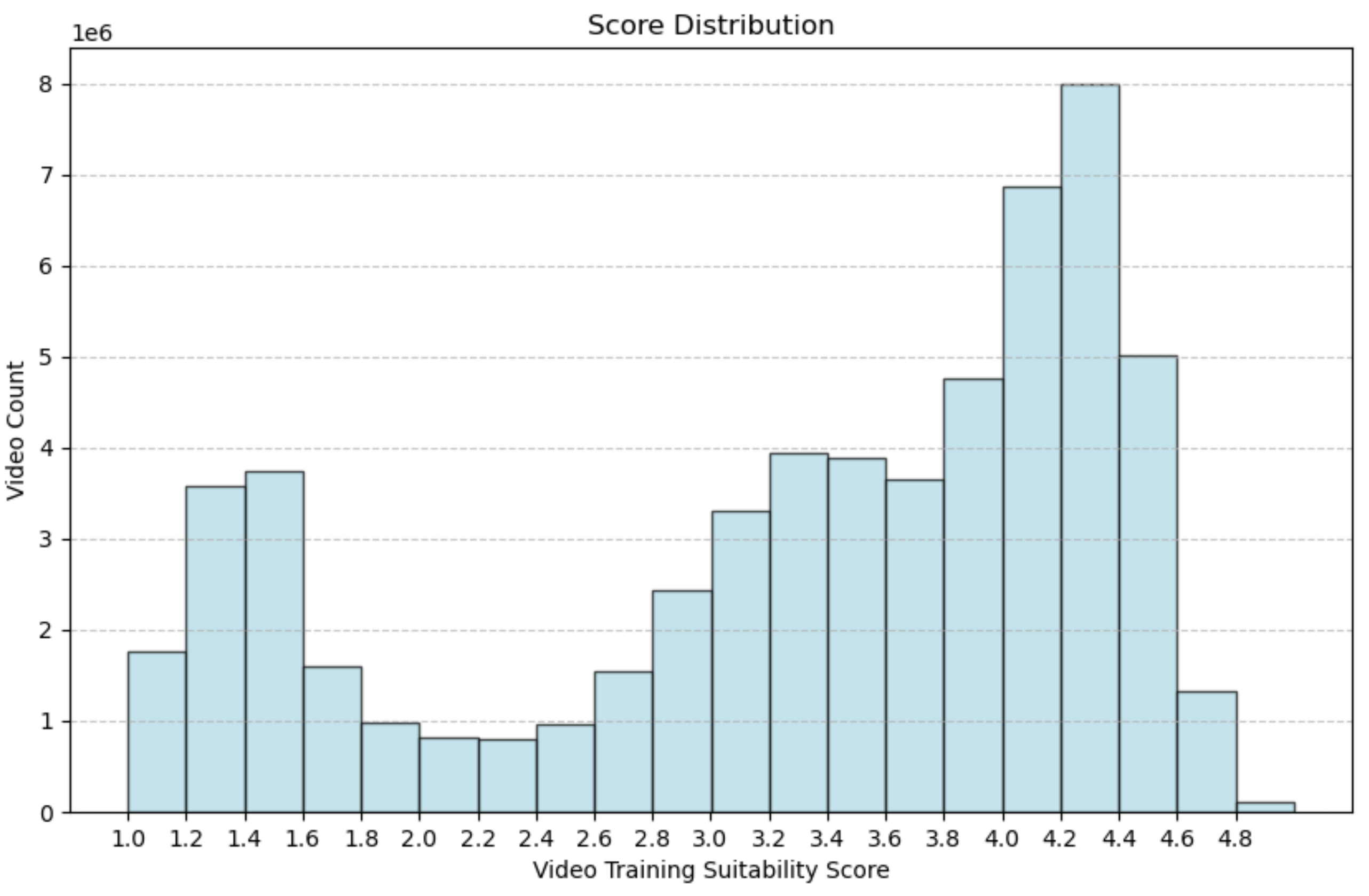}
\end{center}
\vspace{-1.5em}
\caption{\textbf{The distribution of Video Training Suitability Score.}}
\vspace{-1.5em}
\label{fig:VTTS distributin}
\end{figure}

After training \textit{Training Suitability Assessment Network} on the human-aligned dataset, we employ it to predict \textit{Video Training Suitability Score (VTSS)} for all videos, and obtain the score distribution as shown in Fig.~\ref{fig:VTTS distributin}. Since the VTSS distribution can roughly be divided into two Gaussian distributions, we simply chose the decomposition value 2.5 as the VTSS threshold. Based on this threshold, we filtered out a dataset containing a total of 36 million video clips with corresponding captions. We designate this dataset as \DatasetName, which is the final dataset we are presenting.

\vspace{-0.3em}
\subsection{Metrics conditioning}
\vspace{-0.3em}
\label{section: comprehensive perception conditions}

In previous pipelines, data metrics are simply used for data filtering. Meanwhile, the quality of the filtered data still varied, making it difficult for the model to distinguish between high-quality and low-quality data. To address this issue, we propose a more fine-grained conditioning method to incorporate quality information of different videos into the generation model during  the training phase, leading to better consistency between conditions and video content. During the inference stage, this method also enables fine-grained control over the generated videos.

Specifically, during video diffusion training, we first encode data metrics such as motion score, aesthetic score, and clarity score into frequency embeddings. Subsequently, frequency embeddings are passed through an MLP to obtain multiple embeddings, which are then directly added to the timestep embeddings and incorporated into the transformer block using Adaptive Layer Normalization (AdaLN). This method offers  two main advantages. First, it does not increase the computational load of the diffusion model. Second, compared to adding conditions in captions like Open-sora~\citep{opensora}, it supports more precise control with higher sensitive to numerical scores, and posses a stronger ability to decouple control over different metrics. During the inference, we can set different feature scores, such as setting all scores to the highest value, to generate high-quality videos.

\begin{figure}[h]
\vspace{-0.8em}
\begin{center}
\includegraphics[scale=0.23]{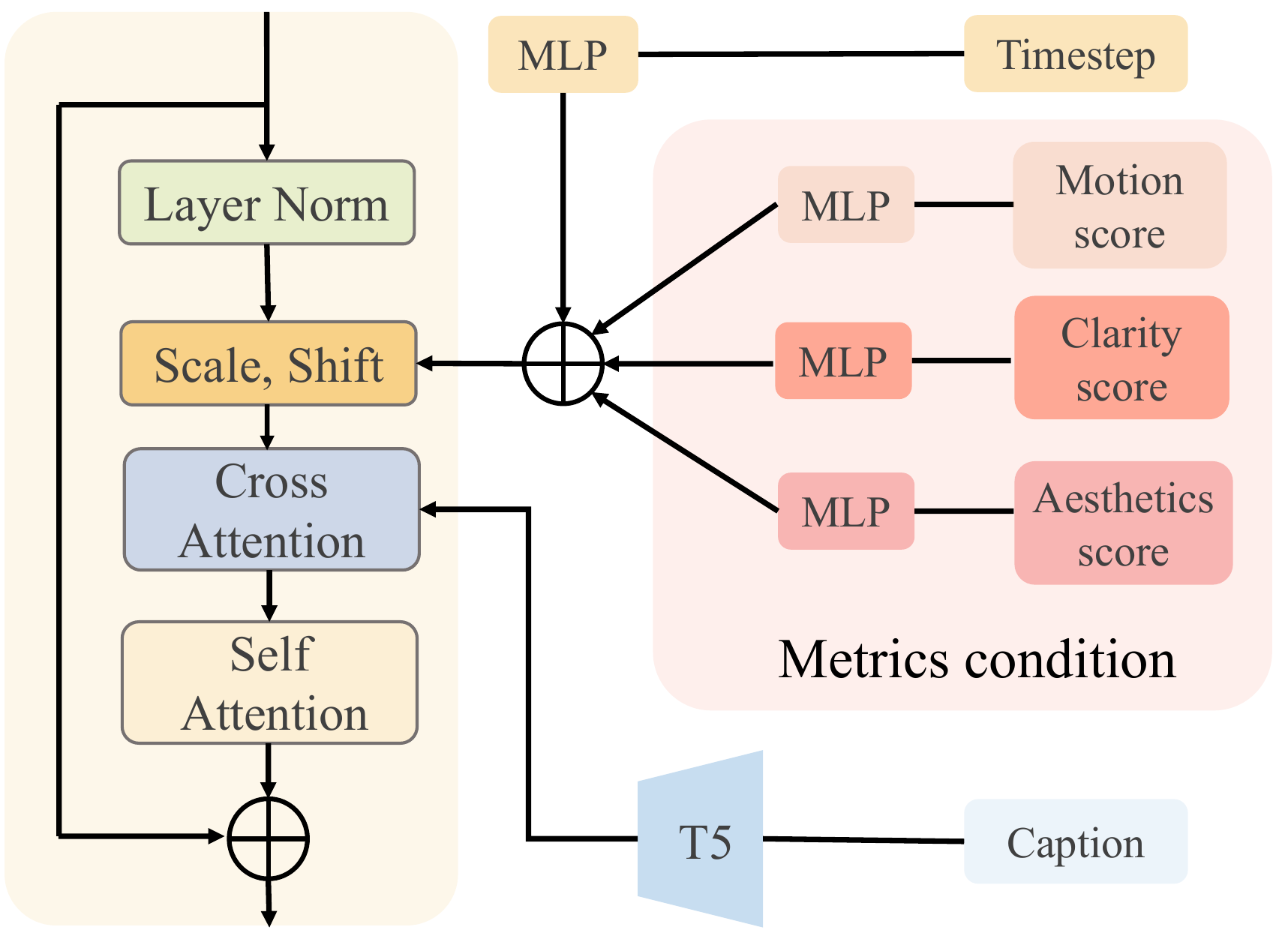}
\end{center}
\vspace{-1.5em}
\caption{\textbf{The pipeline of metrics conditions.}}
\vspace{-1.2em}
\label{fig:fine-grained condition pipeline}
\end{figure}
% tab
\begin{table*}[ht]
\centering
\vspace{-2em}
\caption{\textbf{Quantitative results of text-to-video generation.} We compare the performance of generation models trained on different datasets with VBench. The generation model trained on \DatasetName surpasses other models on both \textbf{quality score} and \textbf{semantic score}, with the highest \textbf{total score}.}
\vspace{-0.5em}
\resizebox{\textwidth}{!}{%
\begin{tabular}{lcccccccccccccccccc}
\toprule
VBench& \makecell{Aesthetic\\Quality} & Scene & \makecell{Subject\\Consistency} & \makecell{Background\\Consistency} & \makecell{Temporal\\Flickering} & \makecell{Motion\\Smoothness} & \makecell{Dynamic\\Degree} & \makecell{Imaging\\Quality} & \makecell{Object\\Class}& \makecell{Multiple\\Objects} \\
\midrule
Panda-70M & 0.3988 & 0.1106 & 0.8584 & 0.9435 & 0.9576 & 0.9742 & 0.7722 & 0.4250 & 0.3017 & 0.0223 \\
Koala-w/o TSA & 0.4808 & 0.2105 & \textbf{0.9335} & \textbf{0.9668} & \textbf{0.9857} & 0.9855 & 0.4222 & 0.5535 & 0.5453 & 0.1154 \\
Koala-37M-manual & 0.4683 & 0.2135 & 0.9388 & 0.9664 & 0.9810 & \textbf{0.9870} & 0.4028 & 0.5422 & 0.4858 & 0.1099 \\
Koala-36M & 0.4832 & 0.1994 & 0.9245 & 0.9613 & 0.9766 & 0.9851 & 0.5750 & \textbf{0.5585} & 0.4739 & 0.1145 \\
Koala-w/o TSA (condition) & 0.5272 & \textbf{0.3211} & 0.9162 & 0.9514 & 0.9210 & 0.9718 & \textbf{0.9833} & 0.5316 & \textbf{0.7734} & 0.2492 \\
Koala-36M (condition)& \textbf{0.5318} & 0.3163 & 0.9222 & 0.9554 & 0.9246 & 0.9768 & 0.9194 & 0.5344 & 0.7794 & \textbf{0.2953} \\
\midrule
VBench & \makecell{Human\\Action} & Color & \makecell{Spatial\\Relationship} & \makecell{Temporal\\Style} & \makecell{Appearance\\Style} & \makecell{Overall\\Consistency} & \makecell{ \textbf{Quality}\\\textbf{Score}} & \makecell{\textbf{Semantic}\\\textbf{Score}} & \makecell{\textbf{Total}\\\textbf{Score}} \\
\midrule
panda-70M & 0.2400 & 0.5942 & 0.0482 & 0.1281 & 0.2014 & 0.1404 & 0.7343 & 0.3093 & 0.6493 \\
Koala-w/o TSA & 0.5180 & 0.8958 & 0.2168 & 0.1630 & 0.1971 & 0.1881 & 0.7758 & 0.4668 & 0.7140 \\
Koala-37M-manual & 0.4700 & 0.9128 & 0.1978 & 0.1589 & 0.2003 & 0.1893 & 0.7704 & 0.4548 & 0.7073 \\
Koala-36M & 0.4880 & \textbf{0.9172} & 0.1923 & 0.1571 & 0.1960 & 0.1850 & 0.7819 & 0.4504 & 0.7156 \\
Koala-w/o TSA (condition)& \textbf{0.8280} & 0.9106 & 0.2434 & 0.2039 & \textbf{0.2019} & 0.2277 & 0.7823 & 0.5874 & 0.7433 \\
Koala-36M (condition)& 0.8080 & 0.8960 & \textbf{0.2689} & \textbf{0.2045} & 0.2009 & \textbf{0.2279} & \textbf{0.7846} & \textbf{0.5915} & \textbf{0.7460} \\
\bottomrule
\end{tabular}%
}
\vspace{-1.0em}
\label{tab:generation results}
\end{table*}

\vspace{-0.4em}
\section{Experiments}
\vspace{-0.4em}
\subsection{Experiment Setting}
\vspace{-0.4em}
To validate he superiority of Koala-36M dataset and the effectiveness of our data processing pipeline, we train the same generation model from scratch on different datasets for comparison. Our text-to-video base model is based on a  Sora-like structure~\citep{sora} with 3D-full attention transformer block. And each basic transformer block includes 2D self-attention, 3D self-attention, and text cross-attention. We use T5 for text embedding and 3D causal VAE for video compression. Since the training was done from scratch, we set the video duration to 2 seconds and the resolution to 256x256 for faster convergence. We train models on 80G A100 GPUs with a batch size of 32 and a learning rate of 0.0001. All models are trained on their respective datasets passing through 140M data samples in total. To evaluate the performance of generation models, we conduct a comprehensive evaluation on the public benchmark VBench~\citep{vbench}. Due to the domain gap between the captions provided by VBench and training set, we performed prompt expansion on the captions in VBench.

\vspace{-0.5em}
\subsection{Quantitative Results}
\begin{figure}[h]
\begin{center}
\includegraphics[scale=0.2]{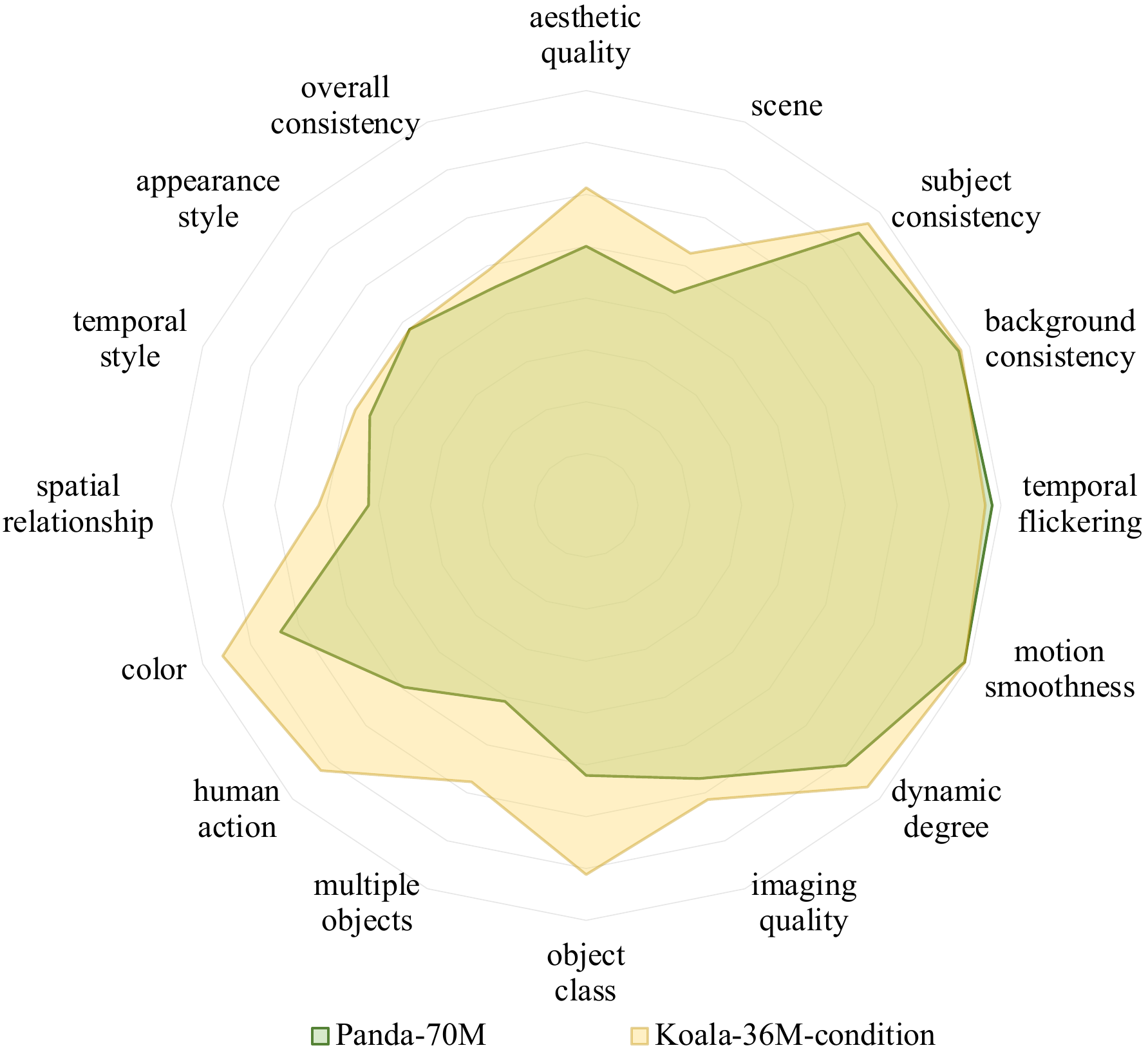}
\end{center}
\vspace{-1em}
\caption{\textbf{Visualization of quantitative results of text-to-video generation.} \DatasetName significantly improves the generation model's performance on aesthetic quality, object class, multi-objects, human action, and color.}
\label{fig:visualize of quantitative results}
\vspace{-1.5em}
\end{figure}

\vspace{-0.5em}
% text
As shown in Tab.~\ref{tab:generation results}, we comprehensively evaluate models trained on Panda-70M and our dataset at the same step. The generation model trained on \DatasetName surpasses other models on both \textbf{quality score} and \textbf{semantic score}, with the highest \textbf{total score}. Furthermore, we visualize the VBench metrics comparison in Fig.~\ref{fig:visualize of quantitative results}. \DatasetName significantly improves the generation model's performance on aesthetic quality, object class, multi-objects, human action, and color.

% figure

\vspace{-0.5em}
\subsection{Qualitative Results}
\begin{figure*}[h]
\begin{center}
\includegraphics[scale=0.22]{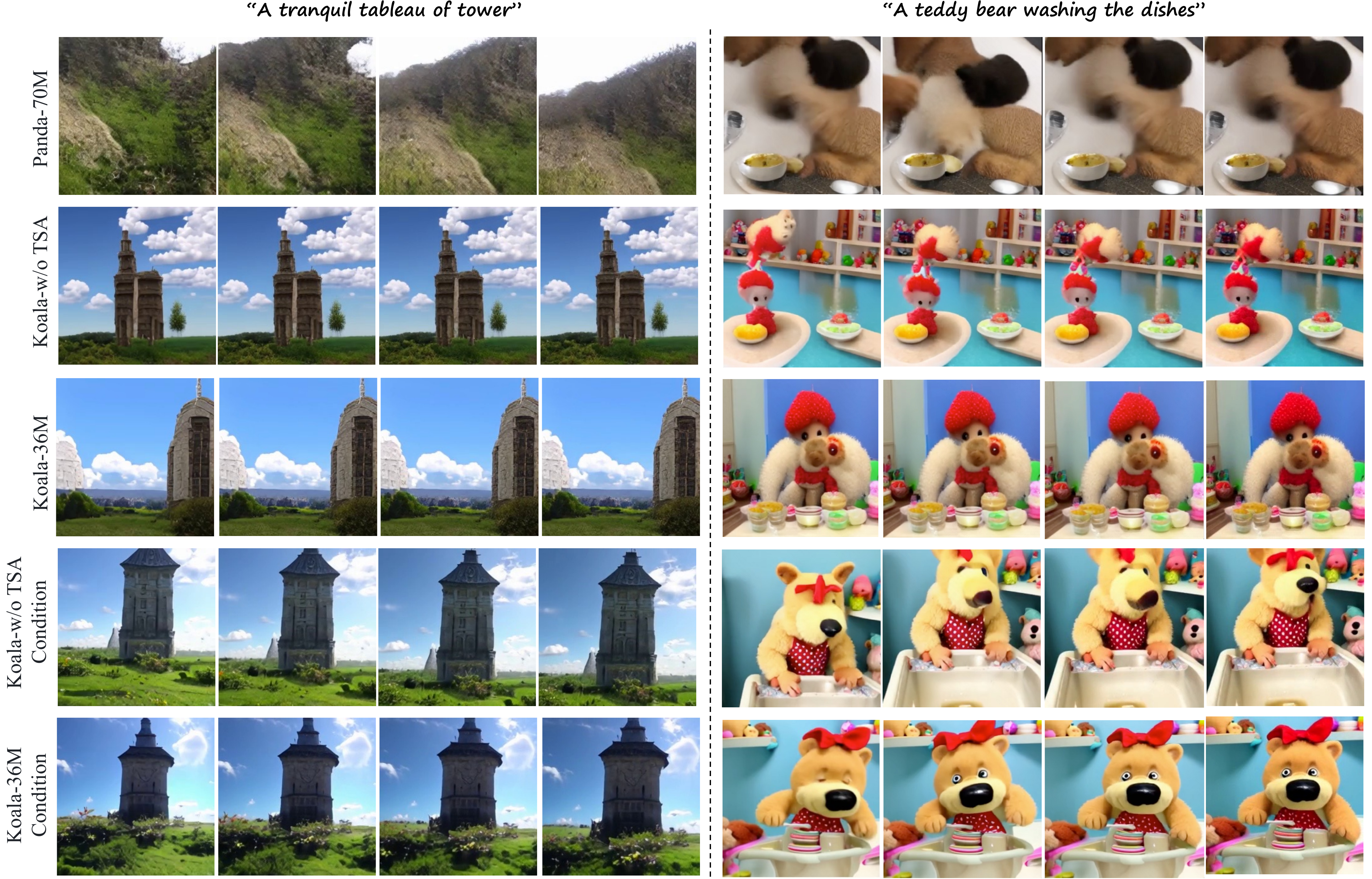}
\vspace{-2em}
\end{center}
\caption{\textbf{Qualitative results of text-to-video generation.} We train the same generation model from scratch on different datasets for comparison. The generation model achieve the optimal performance on \DatasetName, with better video quality and text-video consistency.}
\label{fig:generation results}
\vspace{-1.5em}
\end{figure*}

\vspace{-0.5em}
We visualize the generated videos on VBench's prompts in Fig.~\ref{fig:generation results}. The generation model achieve the optimal performance on Koala-36M , with both the best video quality and text-video consistency. \DatasetName outperform the larger Panda-70M dataset with only 36M data, indicating that our data quality far exceeds that of Panda-70M. See~\ref{app:showcase} for more video generation results.

\vspace{-0.5em}
\subsection{Ablation Experiments}
\vspace{-0.5em}
We conduct extensive ablation experiments to demonstrate the superiority of our dataset and the entire pipeline. Specifically, we performed ablation experiments on different data processing and training strategies, divided into the following groups:
(1) \textbf{Panda-70M}: baseline. 
(2) \textbf{Koala-w/o TSA}: All 48M data without filtering after video splitting and captioning. 
(3) \textbf{Koala-37M-manual}: data filtered from all 48M data by manually multiple thresholds instead of VTSS. 
(4) \textbf{Koala-36M}: filtered dataset from Koala-w/o TSA using VTSS. 
(5) \textbf{Koala-w/o TSA-condition}: All 48M data without filtering but with metrics conditions. 
(6) \textbf{Koala-36M-condition}: Koala-36M with metrics conditions.

\textbf{Data Processing.}
Comparing the results of training from Panda-70M and Koala-w/o TSA in Tab.~\ref{tab:generation results} and Fig.~\ref{fig:generation results}, we find that Koala-w/o TSA produce better results, especially in temporal quality, such as subject consistency, background consistency and temporal flickering. This indicates that our newly proposed re-splitting algorithm can more accurately segment transitions, reducing semantic inconsistencies between video segments. Additionally, our recaptioning algorithm provided more detailed video descriptions, making it easier for the model to learn the relationship between visual and textual information. To further demonstrate the superiority of our splitting and captioning methods, we conducted extensive experiments in the App.~\ref{app:effectiveness of video splitting methods}.

\textbf{Data Filtering.}
Comparing the results of training from Koala-w/o TSA and Koala-36M, Koala-w/o TSA-condition and Koala-36M-condition, we find that the results from the latter one perform better than that from the former datasets. This indicates that filtering out low-quality data and retaining high-quality data are necessary to prevent the model from learning biased distributions from low-quality data. In addition, comparing the results of training from Koala-37M-manual and Koala-36M, it can be concluded that our filtering method based on single VTSS results in better filtering performance, when more high-quality data and less low-quality data being retained. Extensive experiments of Training Suitability Assessment Network are in App.~\ref{app:vqa}.

\textbf{Metrics conditions.}
Comparing the results of training from Koala-36M and Koala-36M-condition, the generation model shows significant improvements in video quality, when metrics conditions are injected into it. This indicates that guiding model training using sub-metrics is necessary, as it helps the model implicitly model the importance of different data. In addition, we compare our AdaLN-based injection method with text-encoder based method~\citep{opensora} in App.~\ref{app:different metrics condition} Fig.~\ref{fig:comparison of metrics conditions}. It can be discovered that our injection method has more precise control and stronger ability to decouple control over different metrics, when the style of videos transfer with the motion score.

\vspace{-0.8em}
\section{Conclusion}
\vspace{-0.6em}

In this paper, we present a large-scale high-quality dataset called \DatasetName, with accurate video splitting, detailed captions and higher quality video content. \DatasetName dataset is currently the only video dataset that simultaneously possesses a large number of videos (over 10M) and high-quality fine-grained text captions (longer than 200 words), significantly improving the quality of large scale video datasets. Additionally, we propose a refined data processing pipeline to further improve the consistency between fine-grained conditions and video content, including better transition detection method, structured caption system, and data filtering method and fine-grained conditioning.

\textbf{Limitations.} Despite all the strength above, \DatasetName is still insufficient to support the training of an extremely large video generation model with over 1B parameters. A larger-scale datasets need to be further collected and processed, which is remained as the future work.

{
    \small
    \bibliographystyle{ieeenat_fullname}
    \bibliography{main}
}

\appendix
\newpage
\appendix
\maketitlesupplementary

\section{Effectiveness of video splitting methods}
\label{app:effectiveness of video splitting methods}

To validate the accuracy and efficiency of our proposed Color-Struct SVM (CSS) for scene transition detection, we conduct the following experiments. We annotate transitions in 10,000 video clips, creating a test set (approximately half of the videos contain transitions). We then apply our proposed method and open-source methods to detect transitions in the test set, recording the precision and recall of the detections. The open-source method is primarily based on Pyscenedetect~\citep{pyscene}, and we test two versions: one that detects transitions based solely on HSL (Hue, Saturation, Lightness) and another that uses both HSL and edge detection. As shown in the Tab.~\ref{tab:transtion}, it can be observed that our transition detection algorithm outperforms the two pyscenedetect-based methods in terms of both precision and recall. Notably, our algorithm achieves a high recall rate, indicating that it rarely misses transitions in videos.
\begin{table}[ht]
\centering
\caption{\bf{Transitions Detection Metrics for Different Methods}}
\tiny
\label{tab:transtion}
\resizebox{\linewidth}{!}{%
\begin{tabular}{lccc}
\toprule
Method & Accuracy & Recall & Precision \\
\midrule
Pydetect(hsl) & 0.4421 & 0.3096 & 0.5920 \\
Pydetect(hsl+edge) & 0.4574 & 0.4146 & 0.5854 \\
Ours & \bf{0.7741} & \bf{0.9395} & \bf{0.7547} \\
\bottomrule
\end{tabular}
}
\end{table}

\begin{figure}[ht]
\begin{center}
\includegraphics[width=\linewidth]{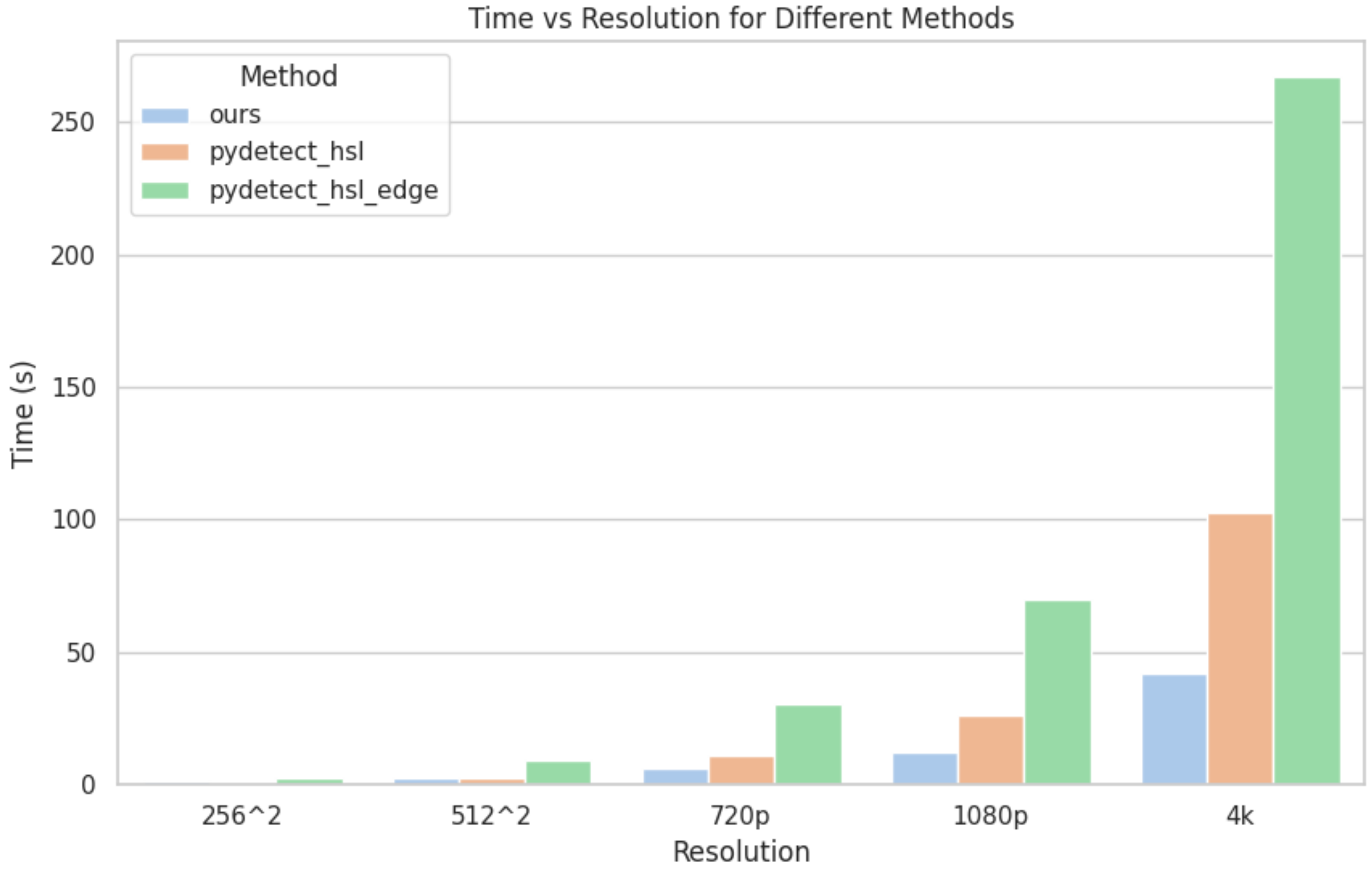}
\end{center}
\vspace{-2em}
\caption{\textbf{Time Consumption for Different Resolutions.} Our method is faster than the others at higher video resolutions.}
\vspace{-1em}
\label{fig:time consumption}
\end{figure}

On the other hand, we compare the runtime efficiency of our method with that of the open-source algorithms. We record the CPU runtime of our algorithm and other open-source algorithms at different resolutions, with the experimental results shown in Tab.~\ref{tab:time_consumption}. We find that our method performs comparably to other methods at 256 resolution. However, as the video resolution increases, our method becomes significantly faster than the other methods (Fig.~\ref{fig:time consumption}).

\begin{table}[ht]
\centering
\caption{\bf{Time Consumption for Different Resolutions and Methods(ms)}}
\label{tab:time_consumption}
\tiny
\resizebox{\linewidth}{!}{%
    \begin{tabular}{lcccc}
    \toprule
    Resolution & Our Method &  \makecell{Pydetect\\(hsl)} & \makecell{Pydetect\\(hsl+edge)}
    \\
    \midrule
    256 & 1.42 & \bf{0.68} & 2.50 \\
    512 & \bf{2.45} & 2.63 & 8.82 \\
    720p & \bf{6.15} & 10.73 & 30.57 \\
    1080p & \bf{12.26} & 26.16 & 70.11 \\
    4k & \bf{41.98}& 102.55 & 267.18 \\
    \bottomrule
    \end{tabular}
}
\end{table}

\section{Elimination of Deviations between True Scores and Labeled Scores}

\label{app:label}

\begin{figure*}[h]
\begin{center}
\includegraphics[scale=0.3]{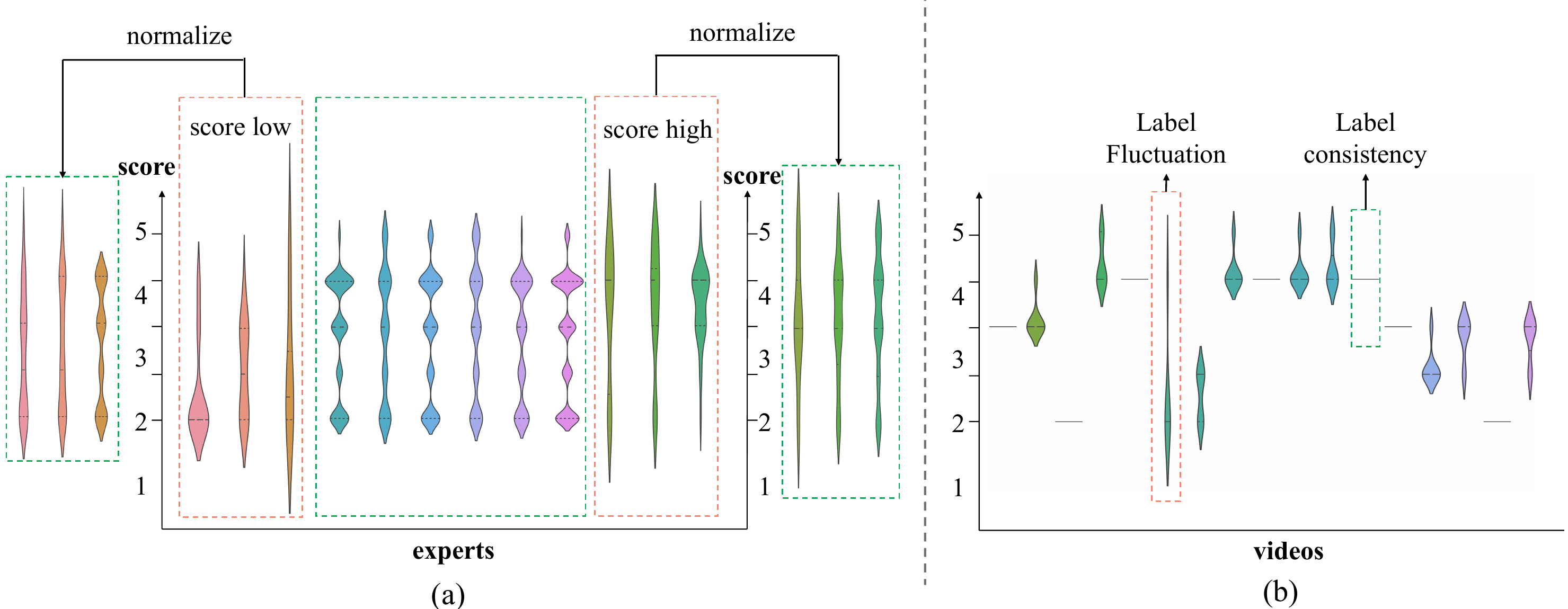}
\end{center}
\vspace{-1em}
\caption{\textbf{Score distribution of different experts and videos.} Fig.(a) visualizes the score distribution of different experts. We eliminate individual preference bias through normalization. Fig.(b) visualizes the score distribution of different videos. We reduce label fluctuation bias with average.}
\vspace{-1.5em}
\label{fig:user distribution of TSA}
\end{figure*}

After establishing the criteria, we randomly sample 200k videos and have it annotated by trained experts, with each video being scored by eight experts on a scale of 1 to 5. To ensure that the annotations closely reflect the true suitability scores, we need to address two types of errors:
\textbf{Individual Preference Bias}: As shown in the Fig.~\ref{fig:user distribution of TSA}(a), we visualize the violin plots of scores given by different experts. The expert on the left tends to give lower scores, while the expert on the right tends to give higher scores. These individual preferences can cause the final scores of some videos to be lower or higher than their actual values. Therefore, we standardize the scores of each expert and then scaled them using the mean and variance of the overall scores to eliminate the bias introduced by different experts. From the figure, it can be seen that the scores processed through our normalization and rescaling methods align more closely with the overall score distribution.
\textbf{Label Fluctuation Bias}: As shown in the Fig.~\ref{fig:user distribution of TSA}(b), each video is annotated by eight experts, and different experts may assign different scores due to varying interpretations of the criteria. This leads to label fluctuations. We use the mean score to reduce the error caused by these fluctuations.

\section{Ablation Study of Training Suitability Assessment Network}
\label{app:vqa}

We conduct comprehensive ablation experiments on our \textit{Training Suitability Assessment Network}. The experimental results are shown in Tab.~\ref{tab:vqa ablation}. The baseline model utilizes only dynamic features. Adding the static branch enables the model to capture more static information, thereby improving overall performance. The inclusion of the feature branch allows the model to leverage additional label information, further enhancing its performance. The WCGB module integrates label information with dynamic and static features through a cross-gating mechanism, achieving optimal performance. Each module addition significantly boosts the model's performance. Combining dynamic and static branches allows the model to capture both types of information. The feature branch utilizes label information for further improvement. The WCGB module optimizes feature integration, achieving the best results. In Tab.~\ref{tab:dover results}, we further supplement the results of Dover to demonstrate the superiority of our TSA module.

% tab
\begin{table}[ht]
\centering
\caption{Performance of Different Combinations}
\vspace{-0.5em}
\label{tab:vqa ablation}
\resizebox{\linewidth}{!}{%
\begin{tabular}{cccccccc}
\toprule
\makecell{Dynamic \\branch
} & \makecell{Static\\branch} & \makecell{Feature\\branch} &WCGB& PLCC$\uparrow$  & SRCC$\uparrow$ & KRCC$\uparrow$ & RMSE$\downarrow$ \\
\midrule
 \checkmark & & & & 0.8684 & 0.8580 & 0.7027 & 0.4644 \\
\checkmark & \checkmark& & & 0.8730 & 0.8637 & 0.7111 & 0.4555 \\
\checkmark& \checkmark&\checkmark & & 0.8953	
 & 0.8864& 	0.7397	& 0.4203  \\
\checkmark& \checkmark&\checkmark &\checkmark & \bf{0.8974} & \bf{0.8868} & \bf{0.7406} & \bf{0.4099} \\
\bottomrule
\end{tabular}
}
\vspace{-1em}
\end{table}

\begin{table}[ht]
\tiny
\centering
\vspace{-1em}
\caption{\textbf{Comparison between Dover and TSA.}}
\vspace{-1em}
\renewcommand{\arraystretch}{0.8}
\resizebox{\linewidth}{!}{%
\begin{tabular}{ccccc}
    \toprule
    Method & PLCC$\uparrow$  & SRCC$\uparrow$ & KRCC$\uparrow$ & RMSE$\downarrow$ \\
    \midrule
    FastVQA & 0.8684 & 0.8580 & 0.7027 & 0.4644 \\
    Dover & 0.8554 & 0.8506 & 0.6788 & 0.6497 \\
    Ours & \bf{0.8974} & \bf{0.8868} & \bf{0.7406} & \bf{0.4099} \\
    \bottomrule
    \end{tabular}
}
\vspace{-1.em}
\label{tab:dover results}
\end{table}

\section{Further Analysis on Data Filtering}
\label{app:data filtering analysis}
We claim existing methods neglect the joint distribution of sub-metrics, resulting in inaccurate thresholds in Sec.~\ref{section: data filtering}. We demonstrate our claim with two conclusions. (1) There exists joint distribution between sub-metrics. As shown in Tab.~\ref{tab:correlation coefficient}, we select three sub-metrics and calculate the pairwise correlation coefficients on the unfiltered data, finding that the different sub-metrics are not independent of each other. (2) Due to the slight deviation of the sub-metrics threshold from the optimal values, the filtering errors accumulate on the errors of each sub-metric, resulting in a larger overall filtering error. As shown in Tab.~\ref{tab:filtering error}, the amount of incorrectly filtered data increases as the number of sub-metrics with inaccurate threshold increases.

% tab
\begin{table}[ht]
\centering
\vspace{-0.5em}
\caption{\textbf{Correlation coefficients between sub-metrics.}}
\vspace{-0.5em}
\resizebox{\linewidth}{!}{%
\begin{tabular}{cccc}
    \toprule
    Correlation coefficients & (Clarity, Aesthetic)  & (Clarity, Motion) & (Motion, Aesthetic) \\
    \midrule
    Pearson correlation & 0.3774 & -0.4028 & -0.2515 \\
    Spearman's rank correlation & 0.3732 & -0.4324 & -0.2347 \\
    \bottomrule
\end{tabular}
}
\vspace{-1.0em}
\label{tab:correlation coefficient}
\end{table}

% tab
\begin{table}[ht]
\centering
\vspace{-0.8em}
\caption{\textbf{The incorrectly filtered data with the increasing number of inaccurate sub-metrics thresholds.}}
\vspace{-0.5em}
\resizebox{\linewidth}{!}{%
\begin{tabular}{cccc}
\toprule
Sub-metrics threshold with deviation (+10\%) & (Clarity)  & (Clarity, Aesthetic) & (Clarity, Motion, Aesthetic) \\
\midrule
Incorrectly filtered data / all data & 250K/48M & 290K/48M & 340K/48M \\ 
\bottomrule
\end{tabular}
}
\vspace{-1.0em}
\label{tab:filtering error}
\end{table}

\vspace{-0.5em}
\section{More Quantitative Results}
As shown in Tab.~\ref{tab:rebuttal quantitative results}, we further pretrain the same model on the HD-VG-130M dataset and evaluate its performance on VBench and additional metrics, such as FVD score. Our dataset outperforms Panda-70M and HD-VG across all metrics. The non-training metrics of other datasets are also presented in Tab. 1 of the main paper. 
Meanwhile, we further conduct experiments on higher resolution (512) and longer duration (4s),  demonstrating the superiority of our dataset.

% tab
\begin{table}[ht]
\centering
\caption{\bf{Quantitative results of text-to-video generation}}
\label{tab:rebuttal quantitative results}
\resizebox{\linewidth}{!}{%
\begin{tabular}{lccccc}
\toprule
 & Quality Score$\uparrow$ & Semantic Score$\uparrow$& Total Score$\uparrow$& FVD$\downarrow$
\\
\midrule
Panda 256-2s      & 0.7343  & 0.3093  & 0.6493 &570.87  \\
HD-VG-130M 256-2s & 0.7696  &  0.4541 & 0.7065 &590.86 \\
Koala-36M (condition) 256-2s  & \textbf{0.7846}  & \textbf{0.5919} & \textbf{0.7460} &\textbf{549.79}   \\
\midrule
Panda 256-4s & 0.7395  &  0.4448 & 0.6806 & 451.09 &  \\
Koala-36M (condition) 256-4s & \textbf{0.7644}  & \textbf{0.4646}  & \textbf{0.7045} &\textbf{354.79}    \\
\midrule
Panda 512-2s & 0.7439  &  0.3954 & 0.6742 & 579.57 &  \\
Koala-36M (condition) 512-2s & \textbf{0.7849}  & \textbf{0.6495}  & \textbf{0.7578}  &\textbf{392.26}   \\
\bottomrule
\end{tabular}
}
\vspace{-0.8em}
\end{table}

\begin{onecolumn}
\section{Comparison of results from different metrics conditions}
\label{app:different metrics condition}

\begin{figure*}[h]
\begin{center}
\includegraphics[width=\textwidth]{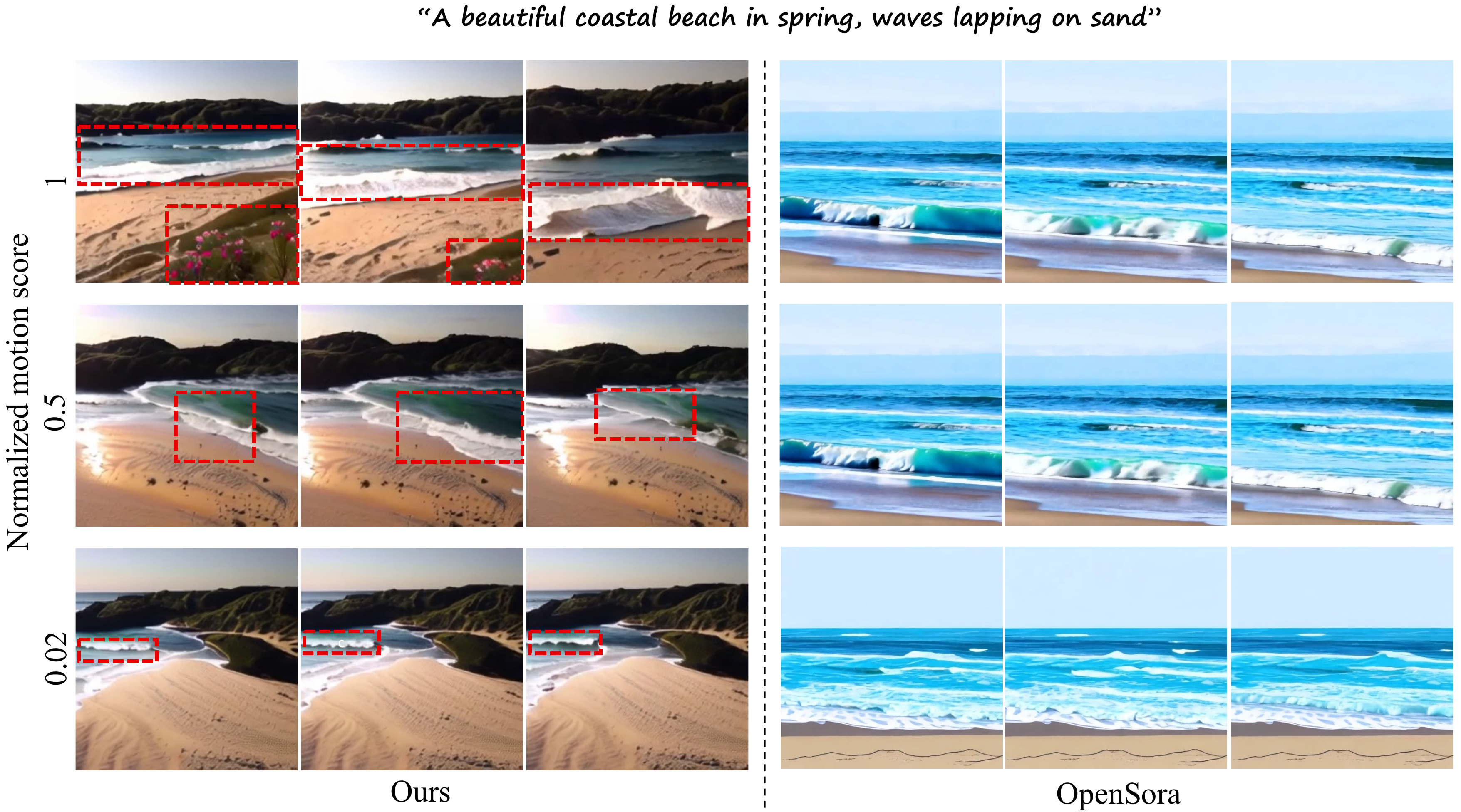}
\end{center}
\caption{\textbf{Comparison of results from different metrics conditions.} Our method has more precise control under the same normalized metrics score and stronger ability to decouple control over different metrics, when the style of videos transfer with the motion score.}
\label{fig:comparison of metrics conditions}
\end{figure*}

\end{onecolumn}

\clearpage
\begin{onecolumn}
\section{More Qualitative Results of Text-to-video Generation}
\label{app:showcase}

\begin{figure*}[h]
\vspace{-1em}
\begin{center}
\includegraphics[scale=0.7]{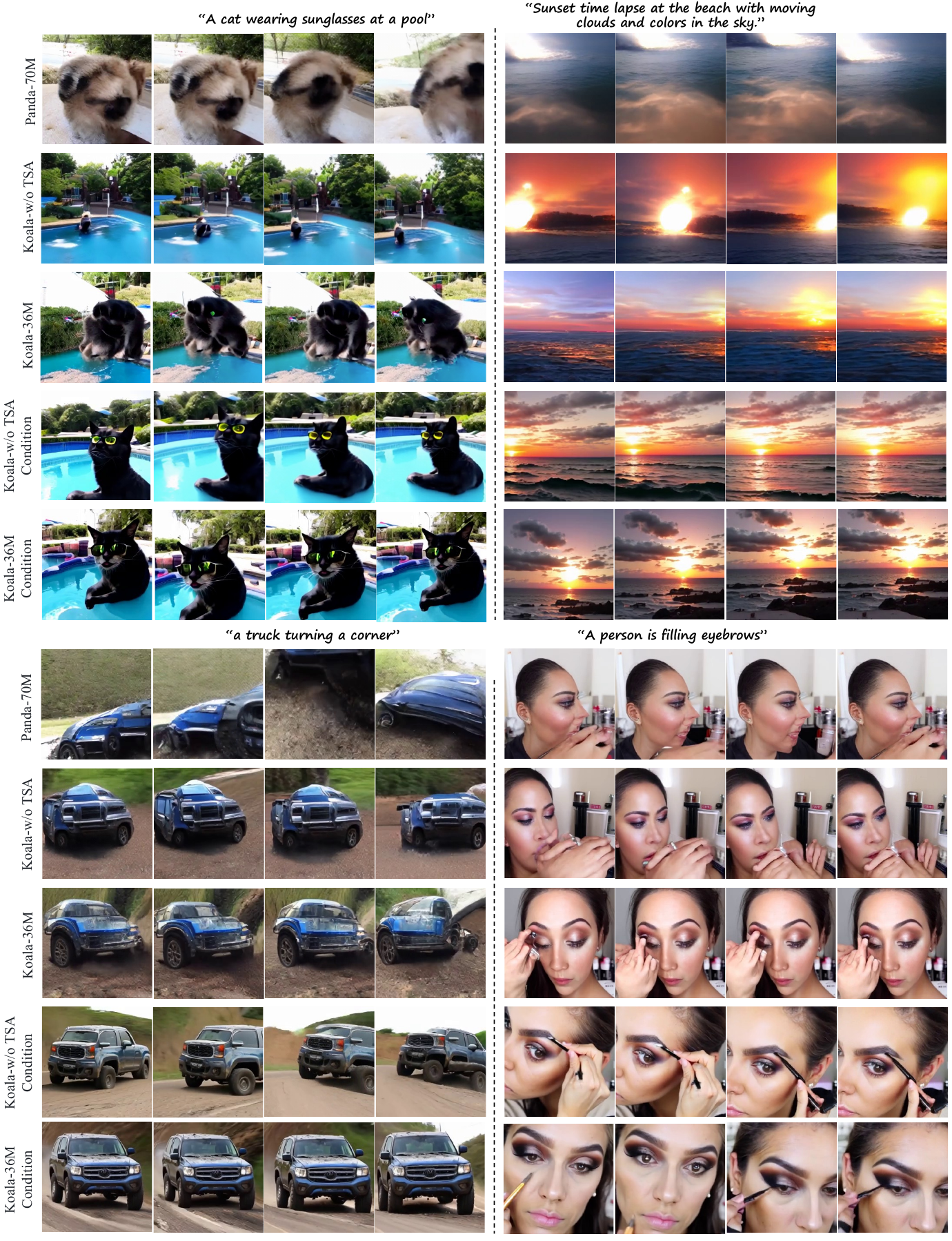}
\end{center}
\vspace{-1.5em}
\caption{\textbf{More qualitative results of text-to-video generation.} }
\label{fig:app:showcase}
\vspace{-1.5em}
\end{figure*}

\end{onecolumn}


\end{document}